\documentclass[11pt]{article}

\usepackage[final]{acl}

\usepackage{times}
\usepackage{latexsym}
\usepackage{algorithm}
\usepackage{algpseudocode}
\usepackage{amsmath}
\usepackage{amssymb}
\usepackage{tcolorbox}
\usepackage[T1]{fontenc}

\usepackage[utf8]{inputenc}

\usepackage{microtype}

\usepackage{inconsolata}

\usepackage{graphicx}
\usepackage{booktabs}
\usepackage{multirow}
\usepackage{makecell}
\tcbuselibrary{breakable}
\newtcolorbox{promptbox}[1]{
  breakable,
  colback=gray!8,
  colframe=black,
  title=#1,
  fonttitle=\normalsize\bfseries,
  fontupper=\normalsize,
  boxrule=0.6pt,
  arc=1pt,
  left=1mm,right=1mm,top=0.6mm,bottom=0.6mm,
  before skip=3pt,after skip=4pt
}
\newtcolorbox{judgepromptbox}[1]{
  breakable,
  colback=gray!8,
  colframe=black,
  title=#1,
  fonttitle=\normalsize\bfseries,
  fontupper=\normalsize,
  boxrule=0.6pt,
  arc=1pt,
  left=0.8mm,right=0.8mm,top=0.4mm,bottom=0.4mm,
  before skip=2pt,after skip=3pt
}
\title{From Retrieval to Reconstruction: Constructing Evolvable Cognitive Memory for Long-Term Dialogue}

\author{
\textbf{Zirui Liao\textsuperscript{1},
Zhengxian Wu\textsuperscript{1},
Zhuohong Chen\textsuperscript{1},
Yunyao Yu\textsuperscript{1}} \\
\textbf{Xiaoyu Liu\textsuperscript{1},
Yifan Xu\textsuperscript{1},
Haoqian Wang\textsuperscript{1}} \\
\textsuperscript{1}Tsinghua University Shenzhen International Graduate School, Shenzhen, China \\
\texttt{liaozr24@gmail.com}
}

\begin{document}
\maketitle

\begin{figure*}[t]
    \centering
    \includegraphics[width=\textwidth]{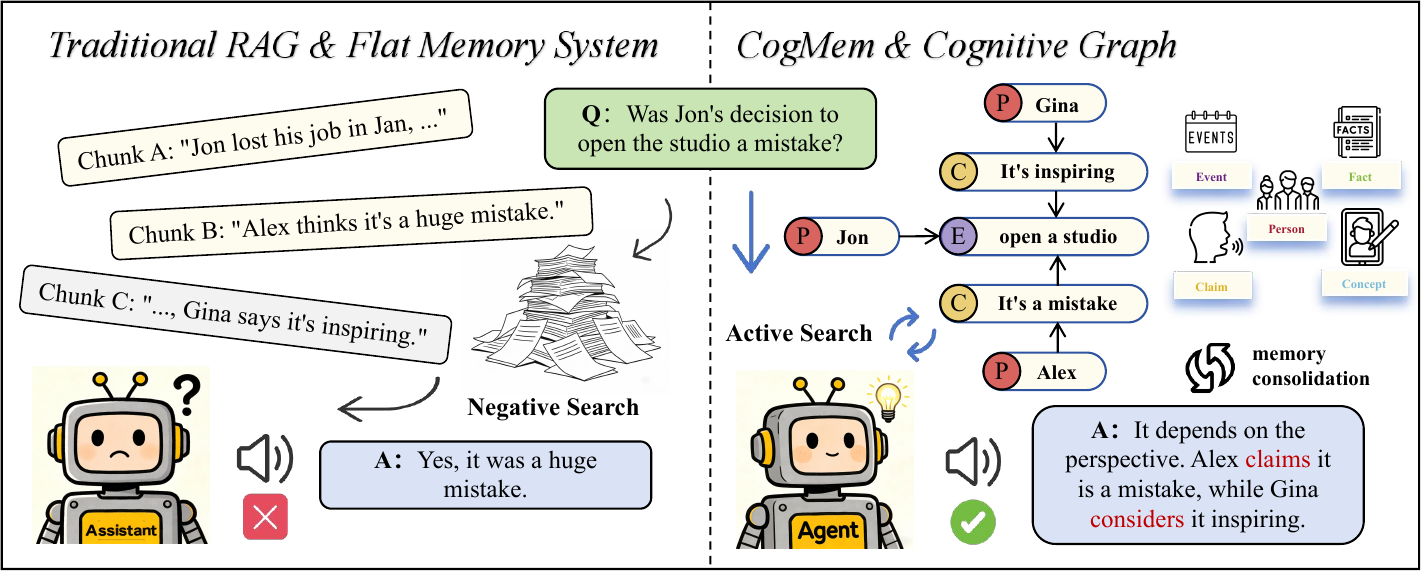}
    \caption{
    \textbf{The Paradigm Shift from Passive Retrieval to Active Reconstruction.} 
    \textit{Left (traditional RAG and flat memory):} passive retrieval over coarse text chunks introduces noise and may miss relevant counter-evidence. Even when a related statement is retrieved, the flat representation obscures its source and can cause the LLM to treat an attributed opinion as an unqualified record.
    \textit{Right (CogMem):} a PEC$^2$F cognitive graph supports active reconstruction, allowing the agent to attribute conflicting viewpoints to their sources and synthesize an evidence-grounded answer.
    }
    \label{fig:teaser}
\end{figure*}

\begin{abstract}
Large Language Models (LLMs) serving as long-term dialogue agents require memory systems that support reliable reasoning over extended interactions. However, existing Retrieval-Augmented Generation (RAG) frameworks typically treat memory as passive storage, making it difficult to distinguish source-attributed beliefs from unattributed event/fact records and to connect evidence dispersed across sessions. We introduce \textbf{CogMem}, a cognitive memory architecture based on the \textbf{PEC$^2$F (Person-Event-Concept-Claim-Fact)} graph schema. Dedicated Claim nodes preserve the source and target of subjective statements, while Fact and Event nodes represent semantic and episodic knowledge. Dialogue turns are incrementally converted into provenance-aware graph records, consolidated into higher-level facts, and reconciled into temporally scoped Claim views when the same source provides conflicting updates. For retrieval, a rule-based controller driven by LLM intent parsing composes four deterministic graph operators---anchoring, traversal, intersection, and evidence grounding---to reconstruct query-relevant context. Experiments on \textbf{LoCoMo} and \textbf{LongMemEval} show strong performance, especially on multi-hop, temporal, and knowledge-update tasks. Ablations and a semantic-collapse probe support complementary contributions from epistemic separation, consolidation, and agentic retrieval. Code: \url{https://github.com/Silent-Rain02/CogMem}.
\end{abstract}

\section{Introduction}
\label{sec:introduction}

The evolution of Large Language Models (LLMs) from stateless text generators to autonomous personal agents has created an urgent need for persistent, high-quality memory \citep{wang2024survey, xi2023rise}. While recent advancements in scaling context windows \citep{gemini15} allow models to buffer vast amounts of recent interaction, extended context alone cannot capture the structured, evolving nature of human biography. True long-term intelligence requires not just storage, but the ability to reconstruct narratives, extract key insights, and understand social contexts of multi-speaker environments \citep{tulving1972episodic}.

Despite this need, current approaches predominantly follow a paradigm of \textbf{passive retrieval}. Standard Retrieval-Augmented Generation (RAG) \citep{lewis2020rag} and early memory systems typically vectorize text chunks into a flat index. We argue that this "store-and-retrieve" paradigm has two key limitations.
First, vector-based methods can suffer from a failure mode that we call \textbf{Semantic Collapse}. In a dense vector space, an event record (e.g., "\textit{Jon lost his job}") and a subjective opinion about it (e.g., "\textit{Gina thinks Jon's job loss is a mistake}") often become very similar due to high semantic overlap. This can cause memory contamination, where an attributed opinion is returned as if it were an unqualified record, undermining Theory-of-Mind reasoning \citep{rabinowitz2018machine}. A controlled probe in Section~\ref{sec:schema} illustrates this failure mode: matched fact--claim pairs have an average BGE-M3 cosine similarity of 0.8231, and flat retrieval misattributes the speaker's opinion as the requested record in 64 of 150 cases (42.7\%).
Second, static retrieval pipelines lack \textbf{intentionality}. Complex dialogue reasoning often requires connecting information across time. A rigid retrieval algorithm cannot dynamically adjust its search strategy based on intermediate results, failing to perform the multi-hop reasoning required for deep understanding \citep{press2023measuring}.

To address these issues, we propose \textbf{CogMem}, a schema-aware cognitive architecture that shifts the design philosophy from passive retrieval to \textbf{agentic active recall}.
Inspired by the Complementary Learning Systems (CLS) theory \citep{kumaran2016learning}, CogMem unifies memory storage and reasoning through a novel \textbf{PEC$^2$F (Person-Event-Concept-Claim-Fact)} schema.
To activate this structure, we implement a \textbf{Cognitive Search Agent} based on the ReAct framework \citep{yao2022react}. Instead of static matching, the agent uses graph operators---such as intersection and temporal scanning---to actively navigate the memory space, reconstructing answers through dynamic reasoning paths.

We evaluate CogMem on the LoCoMo \citep{locomo} and LongMemEval \citep{longmemeval} benchmarks. CogMem outperforms strong baselines, including LightMem \citep{lightmem} and General Agentic Memory \citep{gam2025}, particularly on multi-hop, temporal, and knowledge-update tasks. Our novelty is primarily architectural and representational: CogMem operationalizes Theory of Mind and complementary episodic--semantic learning within a unified, training-free retrieval system rather than proposing an isolated new graph primitive.

In summary, our contributions are threefold:
\begin{itemize}
\item \textbf{Epistemic Completeness through the PEC$^2$F Schema.} We propose a cognitive schema where the \textbf{Attributed--Unattributed Distinction} is modeled through the $C^2$ components: \textit{Concepts} for semantic grounding and \textit{Claims} for epistemic attribution. This structurally addresses the problem of Semantic Collapse.
    
    \item \textbf{Dynamic Stability through Memory Consolidation.} We implement a memory evolution mechanism that dynamically combines sparse episodic traces ($V_E$) into dense semantic knowledge ($V_F$) using an \textbf{Evidence Mounting} strategy. This provides semantic shortcuts while retaining links to granular source evidence, reducing reasoning overhead.
    
\item \textbf{Active Plasticity with Agentic Recall.} We shift the retrieval paradigm from static pipelines to \textbf{agentic active recall}. We design a Cognitive Search Agent equipped with graph operators to navigate the graph, achieving strong performance in multi-hop and temporal reasoning tasks.
\end{itemize}
\section{Related Work}
\label{sec:related_work}

Our work integrates insights from Retrieval-Augmented Generation, Agentic Memory, and Cognitive Science.

\subsection{Retrieval-Augmented Generation (RAG)}
RAG grounds generation in retrieved external context \citep{lewis2020rag, guu2020realm} and can reduce hallucination in knowledge-grounded dialogue \citep{shuster2021retrieval}. Dense and late-interaction retrievers \citep{karpukhin2020dense, khattab2020colbert} are effective for fact-based QA \citep{chen2017reading}, but retrieval quality remains sensitive to whether the required evidence is surfaced, and static ranking does not itself perform compositional reasoning \citep{mallen2023trust, press2023measuring}.

To address this, hybrid approaches combine Knowledge Graphs (KGs) with LLMs \citep{pan2024unifying, yasuanga2021qa}. Recent systems such as LightRAG \citep{lightrag} and HippoRAG 1\&2 \citep{hipporag, hipporag2} use entity--relation graphs or Personalized PageRank to capture dependencies. Their task-agnostic graph structures, however, do not explicitly distinguish source-attributed claims from event and fact records, a distinction required for dialogue reconstruction.

\subsection{Memory Systems for Autonomous Agents}
Recent work focuses on equipping agents with persistent memory.
\textbf{Context Management.} Early approaches like MemGPT \citep{packer2023memgpt} manage long context using operating-system-like paging, while MemoryBank \citep{zhong2024memorybank} uses forgetting curves to update memory.  
\textbf{Structured Memory.} Systems such as Mem0 \citep{mem0} and MemoryOS \citep{memoryos} store user preferences in a hierarchy. A-MEM \citep{amem} proposes a memory bank that improves with experience.
\textbf{Efficiency \& Reasoning.} LightMem \citep{lightmem} uses pruning to reduce latency, while GAM \citep{gam2025} treats recall as an iterative research process.
A separate system also named CogMem uses a three-layer architecture for sustained multi-turn reasoning \citep{zhang2025cogmem}. Our CogMem is distinct: it targets long-term dialogue reconstruction through the PEC$^2$F provenance schema, episodic-to-semantic consolidation, and explicit graph operators such as set intersection.

\subsection{Cognitive Architectures and Theory of Mind}
Cognitive frameworks inspire AI design \citep{sumers2024cognitive}. Generative Agents \citep{park2023generative} demonstrated that agents could simulate realistic behavior by reflecting on memory streams. Our work aligns with the Complementary Learning Systems (CLS) theory \citep{kumaran2016learning, mcclelland1995there}, describing the interplay between fast episodic learning and slow semantic consolidation.
Earlier propositional semantic networks such as SNePS also modeled the intensional belief structures of cognitive agents \citep{shapiro1986sneps}. CogMem shares this emphasis on proposition-level attribution, but targets incremental LLM dialogue ingestion, memory consolidation, and query-adaptive retrieval over long interaction histories.
Crucially, our architecture tackles Theory of Mind (ToM) in LLMs \citep{sap2022neural}. While previous works study LLMs for ToM capabilities, CogMem provides architectural support by explicitly separating source-attributed claims from unattributed event/fact records.
\section{Methodology}
\label{sec:methodology}

\begin{figure*}[t]
    \centering
    \includegraphics[width=1.0\textwidth]{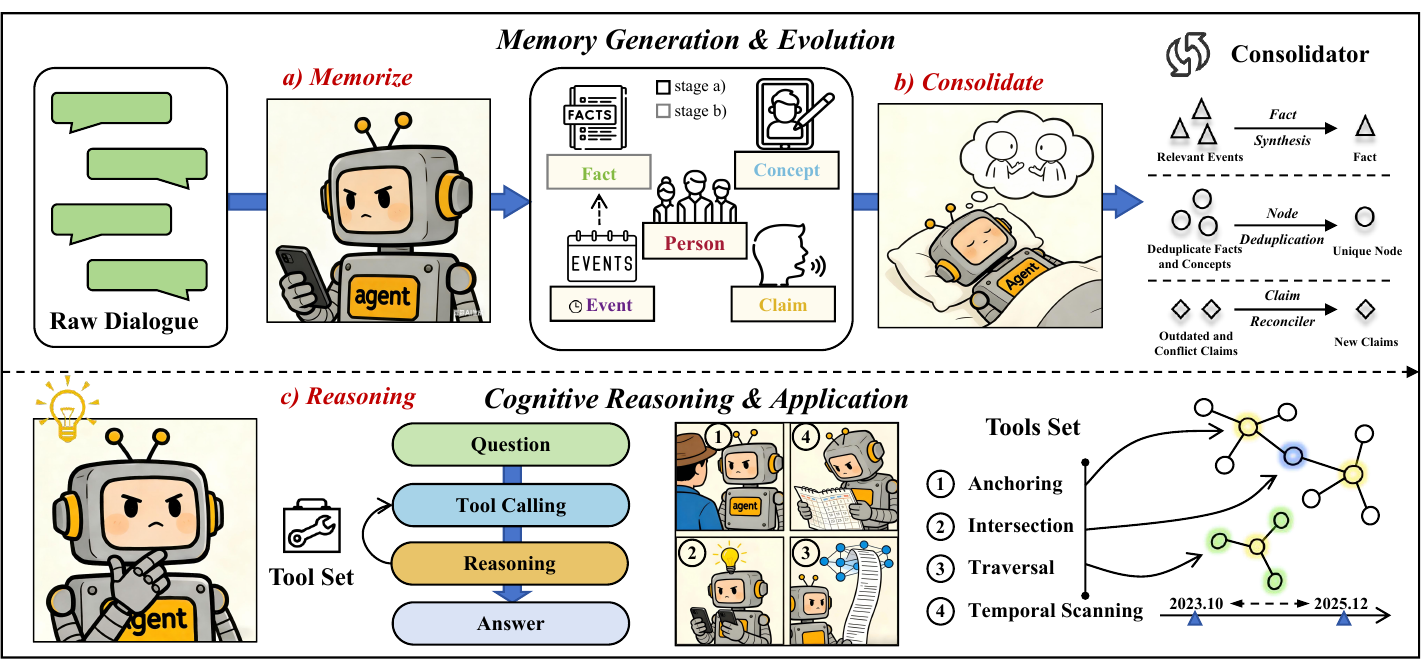}
    \caption{
    \textbf{The Overall Architecture of CogMem.} 
    The framework consists of two layers:
    (1) \textbf{Memory Generation \& Evolution (Top)}: 
    (a) \textit{Memorize}: The system processes raw dialogue to build an initial cognitive graph, converting episodic traces into graph nodes.
    (b) \textit{Consolidate}: An offline mechanism synthesizes Facts from Events, merges duplicate Facts and Concepts, and reconciles temporally conflicting Claims into provenance-linked views without collapsing disagreements across speakers.
    (2) \textbf{Cognitive Reasoning \& Application (Bottom)}:
    (c) \textit{Reasoning}: Instead of passive retrieval, the Search Agent actively recalls information using a specialized \textbf{Tool Set} (Anchoring, Intersection, Traversal, Evidence Grounding) to dynamically navigate the graph and reconstruct answers through a ReAct loop.
    }
    \label{fig:architecture}
\end{figure*}

We propose CogMem, a cognitive architecture that reconstructs long-term memory from continuous dialogue. Unlike traditional RAG systems that focus primarily on storage, we argue that an effective long-term memory system requires three key properties: 
(1) \textit{Epistemic Completeness}---the ability to distinguish source-attributed beliefs from unattributed event/fact records; 
(2) \textit{Active Plasticity}---the capacity to dynamically construct retrieval pathways from vague user intents rather than relying on static indices; and 
(3) \textit{Dynamic Stability}---the mechanism to gradually transform volatile episodic traces into stable semantic knowledge. 
This section describes how CogMem achieves these properties through a unified graph-agent framework.

\subsection{Epistemic Completeness: The \texorpdfstring{PEC$^2$F}{PEC2F} Schema}
\label{sec:schema}

To reduce ambiguity in unstructured memory, we model the memory space as a person-centric graph $\mathcal{G} = (V, E)$ organized by the \textbf{PEC$^2$F} (Person-Event-Concept-Claim-Fact) schema. The vertex set contains five node types: $V = V_P \cup V_E \cup V_C \cup V_{Claim} \cup V_F$.

\paragraph{Problem Formulation: Semantic Collapse.}
A key challenge in dialogue memory is \textit{Semantic Collapse}. Standard dense retrieval systems use an embedding function $\Phi: \mathcal{X} \to \mathbb{R}^d$ that maps both unattributed event/fact records and lexically similar subjective opinions into nearby regions of the vector space. For example, a record $f$ stating "\textit{Jon was fired}" and a subjective claim $c$ stating "\textit{Gina thinks Jon being fired was a mistake}" share most salient content even though they have different epistemic status. A flat retriever can therefore return $c$ when the query asks what happened to Jon.

\paragraph{Semantic-Collapse Probe.}
We directly quantify this failure mode using 150 synthetic fact--claim pairs modeled on conversational patterns in LoCoMo. Each pair contains an unattributed record and a source-attributed opinion with high lexical overlap. With BGE-M3, the mean cosine similarity between paired statements is 0.8231. When queries ask for the event/fact record, a standard flat RAG pipeline selects the subjective claim in 64 of 150 cases (42.7\%). Type-constrained retrieval over the PEC$^2$F graph reduces this error to 9 of 150 cases (6.0\%) by excluding Claim nodes unless the query requests an attributed belief. Table~\ref{tab:semantic-collapse-probe} summarizes the results.

\begin{table}[t]
\centering
\small
\setlength{\tabcolsep}{5pt}
\begin{tabular}{lc}
\toprule
\textbf{Probe statistic} & \textbf{Result} \\
\midrule
\addlinespace[1.5pt]
Mean fact--claim cosine similarity & 0.8231 \\
Flat RAG misattribution & 64/150 (42.7\%) \\
CogMem misattribution & 9/150 (6.0\%) \\
\bottomrule
\end{tabular}
\caption{Direct probe of semantic collapse on 150 matched fact--claim pairs.}
\label{tab:semantic-collapse-probe}
\end{table}

\paragraph{Anchors, Traces, and Facts (The Unattributed Record Spine).}
The graph centers on \textbf{Person} nodes ($V_P$) representing individuals and \textbf{Concept} nodes ($V_C$) representing abstract entities identified through vector clustering. These relatively stable anchors connect to \textbf{Event} nodes ($V_E$), which record specific interactions ($Do$), and \textbf{Fact} nodes ($V_F$), which record observed or self-reported attributes and habits ($Is$). Here, \textit{Fact} is an operational record type rather than a guarantee of independently verified truth. Following Tulving's theory \citep{tulving1972episodic}, an event node $e \in V_E$ links participants to a specific time and place. This design distinguishes recurring patterns (Facts) from unique experiences (Events), providing the temporal information needed for narrative reconstruction.

\paragraph{Separating Beliefs from Facts via Claim Nodes.}
We introduce \textbf{Claim Nodes} ($V_{Claim}$) to model Theory of Mind (ToM) and address Semantic Collapse. Rather than storing a subjective opinion as a direct attribute, we represent it as an intermediate node structure. A statement like "\textit{Jon thinks Gina is lazy}" becomes a two-hop path:
\begin{equation}
    \begin{split}
        (p_{jon}, \texttt{CLAIMS}, n_{claim}) \in E \quad \land \\
        (n_{claim}, \texttt{ABOUT}, p_{gina}) \in E
    \end{split}
    \label{eq:claim}
\end{equation}
where $n_{claim} \in V_{Claim}$ contains the content and sentiment. This additional hop structurally separates a subjective belief from an unattributed record. Even when their embeddings are similar, the graph structure allows the system to identify whether information is stored as a source-attributed belief or as an event/fact record, preserving its epistemic status.

\paragraph{Online Memory Ingestion.}
CogMem updates the graph incrementally after each dialogue turn. The ingestion record contains the utterance, speaker, session identifier, and timestamp. An Atomic Extraction Engine (prompt in Appendix~\ref{sec:Prompts}) maps time-bounded actions to Event nodes, stable self-reported attributes to Fact nodes, and opinions about another person or concept to Claim nodes with explicit \texttt{source} and \texttt{target} fields. Person and Concept strings are normalized and matched against existing anchors before new nodes are created. Every extracted node retains the source-turn identifier and timestamp, enabling the Evidence Grounding operator to recover the original utterance. The resulting nodes and typed edges are inserted immediately; periodic consolidation subsequently adds semantic Fact nodes without deleting their episodic evidence.

\subsection{Active Plasticity: Agentic Retrieval via Cognitive Operators}
\label{sec:agent}

While the PEC$^2$F schema provides a structured representation, its graph complexity makes static retrieval (e.g., fixed k-hop expansion) inefficient for queries involving intent and temporality. To address this, we introduce \textbf{Active Plasticity}, implementing a retrieval controller that dynamically selects cognitive operators based on query intent.

\textbf{Agent Control Logic.}
The retrieval process uses a rule-based controller driven by LLM intent parsing. The operators themselves and their high-level routing rules are deterministic, whereas the intent label and evidence-sufficiency judgment are produced by the LLM. At each step $t$, the agent observes the current subgraph $G_t \subseteq \mathcal{G}$ and selects the next operator $a_t \in \mathcal{O}$ according to Algorithm~\ref{alg:controller}. After the mandatory anchoring call, the process terminates when either sufficient evidence is gathered or the maximum number of post-anchoring controller steps $D_{\text{step}}=5$ is reached.

\begin{algorithm}[t]
\small
\caption{Retrieval Controller Logic}
\label{alg:controller}
\begin{algorithmic}[1]
\Require Query $q$, Graph $\mathcal{G}$, Max controller steps $D_{\text{step}}$
\State $G_0 \gets \emptyset$, $t \gets 0$
\State $V_{\text{anc}} \gets \mathcal{O}_{\text{anchor}}(q)$ \Comment{Always first}
\State $G_0 \gets \text{Subgraph}(V_{\text{anc}})$
\While{$t < D_{\text{step}} \land \neg\text{Ready}(q, G_t)$}
    \State $z \gets \text{ParseIntent}(q, G_t)$
    \If{$z \in \{\texttt{COMPARE}, \texttt{COMMON}\}$}
        \State $a_t \gets \mathcal{O}_{\text{intersect}}(V_{\text{anc}})$ \Comment{Comparison}
    \ElsIf{$z \in \{\texttt{BELIEF}, \texttt{OPINION}\}$}
        \State $a_t \gets \mathcal{O}_{\text{traverse}}(V(G_t), R_{\text{claim}}, \mathcal{T})$
    \ElsIf{$z \in \{\texttt{WHEN}, \texttt{TEMPORAL}\}$}
        \State $a_t \gets \mathcal{O}_{\text{traverse}}(V(G_t), R_{\text{all}}, T)$ \Comment{Temporal}
    \ElsIf{$z \in \{\texttt{WHY}, \texttt{VERIFY}, \texttt{DETAIL}\}$}
        \State $a_t \gets \mathcal{O}_{\text{ground}}(V(G_t))$ \Comment{Ground}
    \Else
        \State $a_t \gets \mathcal{O}_{\text{traverse}}(V(G_t), R_{\text{all}}, \mathcal{T})$
    \EndIf
    \State $G_{t+1} \gets G_t \cup \text{Subgraph}(a_t)$, $t \gets t+1$
\EndWhile
\State \Return $\text{Generate}(q, G_t)$
\end{algorithmic}
\end{algorithm}

\textbf{Intent Parsing and Operator Selection.}
The LLM parses query intent into predefined categories that trigger specific operators:
\begin{itemize}
    \item \textbf{Intersection trigger}: Query contains comparison words ("compare", "both", "common", "difference") $\rightarrow$ execute $\mathcal{O}_{\text{intersect}}$ on anchor nodes.
    \item \textbf{Attribution trigger}: Query asks what one person thinks or believes about another entity $\rightarrow$ execute $\mathcal{O}_{\text{traverse}}$ over Claim relations.
    \item \textbf{Temporal trigger}: Query contains time expressions ("when", "last year", "after") $\rightarrow$ execute $\mathcal{O}_{\text{traverse}}$ with temporal gating $T$ extracted by the LLM.
    \item \textbf{Evidence trigger}: Query asks for verification or supporting detail ("did he really", "why", "evidence") $\rightarrow$ execute $\mathcal{O}_{\text{ground}}$ on current nodes.
\end{itemize}
Here, $R_{\text{claim}}$ contains Claim-specific edge types, $R_{\text{all}}$ contains the permitted typed relations, and $\mathcal{T}$ denotes an unconstrained time range.

\textbf{The Set of Cognitive Operators.}
We define four complementary operators. Table \ref{tab:hyperparams} lists their configurable parameters.

\begin{table}[h]
\centering
\small
\caption{Hyperparameters for Cognitive Operators and Consolidation}
\label{tab:hyperparams}
\renewcommand{\arraystretch}{1.1}
\begin{tabular}{@{}lp{1.2cm}p{2.6cm}c@{}}
\toprule
\textbf{Comp.} & \textbf{Param.} & \textbf{Description} & \textbf{Value} \\
\midrule
\multirow{2}{*}{Anchoring} 
 & $\alpha$ & Lexical vs. semantic balance (Eq.~\ref{eq:anchor-score}) & 0.3 \\
 & $K_{\text{anchor}}$ & Top-K anchor nodes & 15 \\
\midrule
\multirow{2}{*}{Traversal} 
 & $K_{\text{traverse}}$ & Max neighbors per hop & 15 \\
 & $D_{\text{trav}}$ & Max traversal depth & 2 \\
\midrule
Intersection 
 & $K_{\text{intersect}}$ & Candidates per entity for overlap & 10 \\
\midrule
Controller
 & $D_{\text{step}}$ & Max post-anchoring calls & 5 \\
\midrule
\multirow{2}{*}{Consolidation} 
 & $\tau$ & Event cluster threshold (events $\rightarrow$ fact) & 3 \\
 & $\Delta t_{\text{max}}$ & Time window for clustering (days) & 30 \\
\bottomrule
\end{tabular}
\end{table}

\paragraph{Anchoring Operator ($\mathcal{O}_{\text{anchor}}$)}
Maps query $q$ to entry nodes $V_{\text{start}} \subset V$ using hybrid scoring:
\begin{equation}
    S(v, q) = \alpha \cdot \mathbb{I}_{\text{lex}}(v, q) + (1-\alpha) \cdot \cos(\Phi(v), \Phi(q))
    \label{eq:anchor-score}
\end{equation}
where $\alpha=0.3$ (Table \ref{tab:hyperparams}), $\mathbb{I}_{\text{lex}}$ is fuzzy lexical match, and $\Phi$ is BGE-M3 embedding. Returns top-$K_{\text{anchor}}=15$ nodes.

\paragraph{Spatiotemporal Traversal Operator ($\mathcal{O}_{\text{traverse}}$)}
Filters neighbors $\mathcal{N}(v)$ by edge type $R$ and time window $T$ (extracted by LLM from query):
\begin{equation}
    \begin{split}
        \mathcal{O}_{\text{traverse}}(v,R,T) = \{u \in \mathcal{N}(v) \mid {} \\
        \text{type}(v,u) \in R \land \text{time}(u) \in T\}
    \end{split}
\end{equation}
Maximum results per hop: $K_{\text{traverse}}=15$; max traversal depth: $D_{\text{trav}}=2$ (Table \ref{tab:hyperparams}).

\paragraph{Intersection Operator ($\mathcal{O}_{\text{intersect}}$)}
Computes semantic overlap between neighborhoods of multiple entities:
\begin{equation}
    \mathcal{G}_{\text{common}} = \bigcap_{i=1}^{n} \left( \bigcup_{r \in R_{\text{sem}}} \mathcal{N}_r(p_i) \right)
\end{equation}
where $R_{\text{sem}} = \{\texttt{LIKES}, \texttt{USES}, \texttt{PARTICIPATED\_IN}\}$. Uses $K_{\text{intersect}}=10$ candidates per entity (Table \ref{tab:hyperparams}).

\paragraph{Evidence Grounding Operator ($\mathcal{O}_{\text{ground}}$)}
Retrieves source text from raw corpus $\mathcal{D}$:
\begin{equation}
    \mathcal{O}_{\text{ground}}(v) \rightarrow \{ d \in \mathcal{D} \mid \operatorname{id}(d) = \text{source}(v) \}
\end{equation}
Validates graph inferences against original context before final answer generation.

\subsection{Dynamic Stability via Memory Consolidation}
\label{sec:consolidation}

Memory is not a static repository but a dynamic system that must balance specific episode details with stable generalized knowledge. Without consolidation, continuous dialogue events steadily increase graph complexity, expanding the search space and slowing retrieval. To achieve \textbf{Dynamic Stability}, we implement a consolidation mechanism inspired by Complementary Learning Systems (CLS) theory \citep{kumaran2016learning}, modeling the transfer from fast-learning episodic traces to slow-learning semantic structures.

\paragraph{From Atomic Records to Fact Nodes.}
The consolidation process operates offline as a graph rewriting function over the atomic records created during online ingestion. Let $\mathcal{S}_{p,c} = \{e_1, \dots, e_k\} \subseteq V_E$ be a cluster of episodic nodes where person $p$ interacts with concept $c$ within a window of at most $\Delta t_{\text{max}}$. When the cluster size reaches the threshold, $|\mathcal{S}_{p,c}| \geq \tau$, the system triggers an abstraction function $\Psi$ parameterized by an LLM to convert the pattern into a high-level \textbf{Fact Node}:
\begin{equation}
    f_{new}, \mathcal{T}_{env} \leftarrow \Psi(\mathcal{S}_{p,c})
\end{equation}
where $f_{new} \in V_F$ represents the synthesized semantic proposition (e.g., "\textit{Jon has a habit of dancing}") and $\mathcal{T}_{env} = [t_{start}, t_{end}]$ denotes the computed \textbf{Temporal Envelope}. This transformation promotes implicit edge patterns into explicit nodes, compressing the semantic space while preserving temporal boundaries.

\paragraph{Evidence Preservation and Multi-Resolution Retrieval.}
A key risk in memory compression is losing granular details. To balance abstraction with precision, CogMem uses an \textbf{Evidence Mounting} strategy. Rather than discarding original episodes after synthesis, we add provenance edges to connect the abstract fact to its source:
\begin{equation}
    E_{prov} = \{ (f_{new}, \texttt{SUPPORTED\_BY}, e_i) \mid \forall e_i \in \mathcal{S}_{p,c} \}
\end{equation}
This hierarchical structure enables the Search Agent to operate at \textbf{two resolutions}. For high-level queries (e.g., "\textit{What are Jon's hobbies?}"), the agent retrieves the consolidated Fact node directly, reducing search cost. When the user asks for specifics (e.g., "\textit{When did he first go?}"), the agent follows the \texttt{SUPPORTED\_BY} edges to access the original episodic layer.

\paragraph{Provenance-Preserving Claim Reconciliation.}
CogMem reconciles Claims only when they share a canonical source, target, and predicate. It orders them temporally and creates a provenance-linked view marked as \texttt{CURRENT}, \texttt{SUPERSEDED}, or \texttt{CONFLICTING}; original Claims remain accessible. Claims from different speakers are never merged, so reconciliation resolves updates without erasing disagreement.

\section{Experiments}
\label{sec:experiments}

We evaluate CogMem to answer two primary research questions: (1) How does the proposed schema-aware agent compare against existing memory and graph-retrieval paradigms? (2) Does agentic active recall generalize better than learned traversal models on complex dialogue tasks?

\begin{table*}[t]
\centering
\small
\caption{Performance comparison on the \textbf{LoCoMo} benchmark. Metrics are F1 Score and BLEU-1. Best results are in \textbf{bold}, and the second best are \underline{underlined}.}
\label{tab:main-locomo}
\renewcommand{\arraystretch}{1.2}
\setlength{\tabcolsep}{5pt}
\begin{tabular}{cl|cccccccc}
\toprule
\multirow{2}{*}{\textbf{Backbone}} & \multirow{2}{*}{\textbf{Method}} & \multicolumn{2}{c}{\textbf{Single Hop}} & \multicolumn{2}{c}{\textbf{Multi Hop}} & \multicolumn{2}{c}{\textbf{Temporal}} & \multicolumn{2}{c}{\textbf{Open Domain}} \\
\cline{3-10}
& & \textbf{F1} & \textbf{BLEU-1} & \textbf{F1} & \textbf{BLEU-1} & \textbf{F1} & \textbf{BLEU-1} & \textbf{F1} & \textbf{BLEU-1} \\
\midrule

\multirow[c]{8}{*}{\rotatebox[origin=c]{90}{\textbf{GPT-4o-mini}}}
& \textsc{Naive RAG} & 52.45 & 47.94 & 27.50 & 20.13 & 46.07 & 40.35 & 23.23 & 17.94 \\
\cline{2-10}
& \textsc{LightRAG} & 42.57 & 33.82 & 28.46 & 23.75 & 22.85 & 16.18 & \underline{54.33} & 49.61 \\
& \textsc{HippoRAG} & 39.81 & 31.19 & 39.79 & \underline{37.40} & 26.74 & 22.31 & 51.41 & \underline{50.15} \\
& \textsc{RoG (Learned)} & 55.30 & 50.20 & \underline{44.15} & 35.80 & 32.40 & 26.50 & 35.60 & 30.10 \\
\cline{2-10}
& \textsc{Mem0} & 47.65 & 38.72 & 38.72 & 27.13 & 48.93 & 40.51 & 28.64 & 21.58 \\
& \textsc{LightMem} & 41.79 & 37.83 & 29.78 & 24.80 & 43.71 & 39.72 & 16.89 & 13.92 \\
& \textsc{GAM} & \underline{57.75} & \underline{52.10} & 42.29 & 34.44 & \underline{59.45} & \underline{53.11} & 33.30 & 26.97 \\
\cline{2-10}
& \textbf{\textsc{CogMem (Ours)}} & \textbf{59.20} & \textbf{56.40} & \textbf{50.70} & \textbf{46.10} & \textbf{63.80} & \textbf{57.90} & \textbf{56.10} & \textbf{51.20} \\
\midrule

\multirow[c]{8}{*}{\rotatebox[origin=c]{90}{\textbf{Qwen2.5-14B}}}
& \textsc{Naive RAG} & 47.87 & 42.79 & 26.38 & 19.54 & 30.78 & 25.97 & 14.16 & 10.52 \\
\cline{2-10}
& \textsc{LightRAG} & 41.50 & 34.42 & 26.03 & 21.73 & 20.86 & 16.83 & \underline{54.03} & 48.65 \\
& \textsc{HippoRAG} & 33.10 & 25.74 & 34.78 & 31.15 & 24.88 & 18.56 & 53.42 & \underline{48.98} \\
& \textsc{RoG (Learned)} & 52.40 & 47.10 & 41.80 & 33.20 & 28.60 & 23.10 & 32.50 & 28.40 \\
\cline{2-10}
& \textsc{A-MEM} & 33.75 & 30.04 & 22.09 & 15.28 & 27.19 & 22.05 & 13.49 & 10.74 \\
& \textsc{Mem0} & 42.58 & 35.15 & 31.73 & 24.82 & 28.96 & 26.24 & 15.03 & 11.28 \\
& \textsc{GAM} & \underline{58.93} & \underline{53.74} & \underline{42.96} & \underline{34.48} & \underline{51.52} & \underline{44.43} & 30.63 & 26.04 \\
\cline{2-10}
& \textbf{\textsc{CogMem (Ours)}} & \textbf{60.40} & \textbf{55.80} & \textbf{48.42} & \textbf{44.65} & \textbf{56.20} & \textbf{49.10} & \textbf{55.80} & \textbf{50.10} \\
\bottomrule
\end{tabular}
\end{table*}

\subsection{Experimental Setup}

\paragraph{Datasets and Evaluation Metrics.}
We evaluate CogMem on two established long-term memory benchmarks: \textbf{LoCoMo} \citep{locomo} and \textbf{LongMemEval} \citep{longmemeval}. LoCoMo contains ten long, multi-session conversations and approximately 1.5K question--answer pairs spanning single-hop, multi-hop, temporal, and open-domain reasoning. LongMemEval contains 500 questions over long cross-session histories, testing temporal reasoning, knowledge updating, abstention, and personalized recall. We report F1 and BLEU-1 for LoCoMo following prior work, and LLM-as-judge accuracy for LongMemEval following its established protocol. Dataset processing and category details are provided in Appendix~\ref{sec:appendix_datasets}.

\paragraph{Baselines.}
We compare CogMem against four categories of memory systems: (1) \textbf{Standard Retrieval}: Naive RAG \citep{lewis2020rag}; (2) \textbf{LLM Memory Systems}: Mem0 \citep{mem0}, A-MEM \citep{amem}, LightMem \citep{lightmem}, GAM \citep{gam2025}; (3) \textbf{Heuristic Graph Retrieval}: LightRAG \citep{lightrag}, HippoRAG \citep{hipporag}; (4) \textbf{Learned Graph Traversal}: RoG \citep{rog2024}. 

To improve comparability, we control shared components where each method permits: 
(i) \textbf{BGE-M3} \citep{chen2024bge} for dense embedding (1024-d) to reduce variation from different representation models;
(ii) \textbf{Identical answer-generation prompts} (Appendix \ref{subsec:answer_prompts}) to isolate retrieval quality from prompt engineering variance;
(iii) \textbf{Comparable context limits}: all methods use the same answer-model context limit, while method-specific retrieval limits follow official or recommended configurations. CogMem uses the hyperparameters in Table \ref{tab:hyperparams}; baseline settings are documented in Appendix \ref{sec:baseline_settings}. Architectural differences prevent exact matching of all intermediate computations.

\paragraph{Implementation Details and Evaluation Robustness.}
Experiments use \textbf{GPT-4o-mini} (closed-source) and \textbf{Qwen2.5-14B-Instruct} (open-source) as backbones. CogMem operates with the parameters from Table \ref{tab:hyperparams}: anchoring retrieves top-15 nodes ($K_{\text{anchor}}=15$), traversal explores up to 2 hops ($D_{\text{trav}}=2$) with max 15 neighbors per hop, and the controller executes at most 5 post-anchoring operator calls ($D_{\text{step}}=5$). Temperature is fixed at 0.0 to reduce sampling variability.

Because BLEU and automated judges provide incomplete views of reasoning quality, we additionally conduct a blind human preference study on 50 randomly sampled LoCoMo Multi-Hop and Temporal queries. Two graduate students specializing in artificial intelligence independently annotate all comparisons using reasoning correctness and answer helpfulness as criteria. Disagreements are adjudicated by a third annotator who selects one of the two primary labels, yielding a majority decision; no monetary compensation is provided. CogMem wins 68\% of comparisons, ties 20\%, and loses 12\%. Appendix~\ref{sec:human_evaluation} details the annotation protocol.

\subsection{Main Results}

\paragraph{Performance on Conversational Reasoning (LoCoMo).}
Table \ref{tab:main-locomo} shows F1 and BLEU-1 scores on LoCoMo. CogMem achieves competitive results across both backbones, with particular strength in tasks requiring structured reasoning.

The largest improvements appear in \textit{Multi-Hop} and \textit{Temporal} categories. With Qwen2.5-14B, CogMem reaches 48.42\% F1 on Multi-Hop queries, exceeding the best memory baseline (GAM, 42.96\%), the learned traversal baseline (RoG, 41.80\%), and the best heuristic graph baseline (HippoRAG, 34.78\%). These results are consistent with the intended role of the intersection and traversal operators: dynamically identifying shared concepts or events that bridge entities and sessions.

In \textit{Open Domain} scenarios, which test broad semantic recall rather than precise graph operations, CogMem also edges the strongest competing methods under both backbones. Thus, the structured PEC$^2$F representation retains broad recall while improving categories that require explicit relational or temporal reasoning.

\paragraph{Long-Term Consistency (LongMemEval).}
Table \ref{tab:main-longmem} reports overall LongMemEval accuracy, measuring the system's ability to track coherent states over time.

CogMem achieves the highest overall accuracy with both backbones. When user states evolve across sessions, retrieval systems can surface outdated and current records together. The controlled ablation in Table~\ref{tab:ablation-components} further isolates this behavior: removing Claim nodes reduces Knowledge Update accuracy from 73.08\% to 56.67\%, supporting explicit epistemic attribution as a key contributor.

\begin{table}[h]
\centering
\small
\caption{Overall Accuracy (\%) on \textbf{LongMemEval}. Best results are in \textbf{bold}, and second-best results are \underline{underlined}.}
\label{tab:main-longmem}
\renewcommand{\arraystretch}{1.15}
\setlength{\tabcolsep}{6pt}
\begin{tabular}{l|cc}
\toprule
\multirow{2}{*}{\textbf{Method}} & \multicolumn{2}{c}{\textbf{Accuracy (\%)}} \\
& \textbf{GPT-4o-mini} & \textbf{Qwen2.5-14B} \\
\midrule
\textsc{Naive RAG} & 61.00 & 60.80 \\
\midrule
\textsc{LightRAG} & 52.93 & 47.63 \\
\textsc{HippoRAG} & 54.34 & 51.61 \\
\textsc{RoG} & 56.10 & 53.40 \\
\midrule
\textsc{A-MEM} & 62.60 & \underline{65.20} \\
\textsc{Mem0} & 53.61 & 39.51 \\
\textsc{LightMem} & \underline{64.29} & 61.95 \\
\textsc{GAM} & 63.82 & 58.71 \\
\midrule
\textbf{\textsc{CogMem}} & \textbf{68.40} & \textbf{66.50} \\
\bottomrule
\end{tabular}
\end{table}

\paragraph{Agentic Search vs. Learned Traversal.}
To assess whether zero-shot agentic reasoning offers advantages over learned graph traversal, we compare CogMem against RoG \citep{rog2024}. RoG trains a language model to generate relation paths over knowledge graphs; we evaluate its pre-trained checkpoint in zero-shot mode on LoCoMo and LongMemEval because these datasets lack training splits for path learning. CogMem is likewise training-free, so this comparison measures out-of-the-box transfer rather than the advantage of task-specific supervision; it should not be interpreted as a definitive comparison to a fully fine-tuned RoG system.

RoG performs competitively on \textit{Single-Hop} queries where path patterns are straightforward, but its accuracy is lower on \textit{Open Domain} and \textit{Temporal} questions (Table \ref{tab:main-locomo}). One plausible explanation is that zero-shot relation-path generation transfers less directly when queries require temporal filtering or adaptive operator selection. In contrast, CogMem maps natural-language intents to atomic operators at inference time. Because RoG is evaluated without task-specific fine-tuning, this result should be interpreted as a comparison of zero-shot transfer rather than a general claim about learned traversal.

\subsection{Ablation Studies}
\label{sec:ablation}

To identify the contributions of the cognitive schema (specifically the Claim nodes), the memory consolidation mechanism, and the agentic search loop, we conduct ablation studies using the Qwen2.5-14B backbone.

\paragraph{Ablation Configurations.}
We examine whether improvements stem from epistemic separation, memory evolution, or agentic navigation by defining three ablated variants: (1) \textbf{CogMem w/o Claim} retains the base graph but removes epistemic separation, flattening source-attributed claims into unattributed attribute edges (e.g., $Person \xrightarrow{IS} Lazy$); (2) \textbf{CogMem w/o Consolidation} disables offline consolidation, forcing retrieval from sparse episodic traces without semantic Fact nodes; (3) \textbf{CogMem w/o Agent} replaces the ReAct agent with a deterministic pipeline (anchoring + fixed 2-hop expansion).

Table \ref{tab:ablation-components} shows results on LoCoMo (Multi-Hop F1) and LongMemEval (Knowledge Update Acc). Each component contributes to a distinct capability in the evaluated settings. Removing the agentic loop causes the largest drop: Multi-Hop F1 falls 25.61 points (48.42\% $\rightarrow$ 22.81\%), indicating that the fixed pipeline is less effective on queries requiring adaptive operator selection. Removing Claim nodes reduces Knowledge Update accuracy by 16.41 points (73.08\% $\rightarrow$ 56.67\%), supporting explicit epistemic separation. Omitting consolidation reduces Multi-Hop F1 to 43.56\% as the agent navigates episodic traces without synthesized semantic shortcuts.

\begin{table}[h]
\centering
\caption{Ablation study: removing consolidation, Claim nodes, and ReAct agent.}
\label{tab:ablation-components}
\renewcommand{\arraystretch}{1.05}
\setlength{\tabcolsep}{4pt}
\begin{tabular}{@{}l c c@{}}
\toprule
& \textbf{F1 (\%)} & \textbf{Acc (\%)} \\
\textbf{Configuration} & \textbf{Multi-Hop} & \textbf{Know.Upd.} \\
\midrule
CogMem (Full) & \textbf{48.42} & \textbf{73.08} \\
w/o Consolidation & 43.56 & 66.67 \\
w/o Claim (Flat) & 41.20 & 56.67 \\
w/o Agent (Fixed) & 22.81 & 48.10 \\
\bottomrule
\end{tabular}
\end{table}

\paragraph{Effectiveness of Memory Consolidation.}
We evaluate the offline consolidation mechanism that groups episodic traces into semantic facts, comparing system performance before and after consolidation.

As shown in Table \ref{tab:ablation-consolidation}, consolidation provides two benefits. First, it improves Multi-Hop F1 by 4.86 percentage points by creating semantic connections (e.g., grouping repeated events into a generalized habit node), reducing the chance that the agent misses scattered evidence. Second, it improves retrieval efficiency. By querying consolidated Fact nodes instead of traversing all episodic Event nodes, the average reasoning steps decrease from 3.42 to 2.67. These results show that consolidation supplies useful semantic shortcuts as the graph grows.

\begin{table}[h]
\centering
\caption{Impact of memory consolidation on accuracy (F1) and efficiency (steps).}
\label{tab:ablation-consolidation}
\renewcommand{\arraystretch}{1.05}
\setlength{\tabcolsep}{2.5pt}
\begin{tabular}{@{}l cc cc@{}}
\toprule
& \multicolumn{2}{c}{\textbf{F1 (\%)}} & \multicolumn{2}{c}{\textbf{Steps}} \\
\cmidrule(lr){2-3} \cmidrule(lr){4-5}
\textbf{Category} & Before & \textbf{After} & Before & \textbf{After} \\
\midrule
Single Hop & 58.06 & \textbf{60.40} & 2.84 & \textbf{2.35} \\
Multi Hop & 43.56 & \textbf{48.42} & 3.42 & \textbf{2.67} \\
Temporal & 56.20 & 56.20 & 2.78 & \textbf{2.44} \\
Open Domain & 54.50 & \textbf{55.80} & 2.86 & \textbf{2.55} \\
\bottomrule
\end{tabular}
\end{table}

\paragraph{Further Analyses.}
Appendices~\ref{sec:appendix_efficiency}, \ref{sec:appendix_sensitivity}, and \ref{sec:appendix_error_analysis} provide detailed analyses of construction cost and graph growth, operator sensitivity, and 100 manually categorized LoCoMo failures. These results show that consolidation creates useful semantic shortcuts at the cost of additional graph storage, while remaining errors primarily arise from embedding drift, incomplete search scope, and temporal normalization.

\section{Conclusion}
\label{sec:conclusion}

We introduced \textbf{CogMem}, a training-free architecture that shifts long-term dialogue memory from passive retrieval toward active reconstruction. Its PEC$^2$F schema separates source-attributed Claims from Event and Fact records, periodic consolidation links episodic evidence to reusable semantic summaries, and a query-adaptive controller composes deterministic graph operators for retrieval. Across LoCoMo and LongMemEval, CogMem performs particularly well on multi-hop, temporal, and knowledge-update settings. The component ablations and semantic-collapse probe further show that epistemic separation, consolidation, and agentic navigation contribute complementary capabilities. Overall, the results suggest that reliable long-term memory depends not only on storing more context, but also on preserving provenance and reconstructing evidence through task-appropriate retrieval paths.

The results also expose practical trade-offs. Additive consolidation increases graph storage while reducing online reasoning steps, and the remaining failures concentrate on embedding drift, incomplete search scope, and temporal normalization. CogMem should therefore be viewed not as eliminating retrieval error, but as making long-term memory more structured, attributable, and auditable. Future work can build on this foundation through hybrid retrieval, incremental consolidation, and more adaptive operator selection.

\section{Limitations}
\label{sec:limitations}

\textbf{Empirical Scope.}
Our evaluation covers two English long-term dialogue benchmarks and two LLM backbones; it does not establish generalization to multilingual dialogue, noisy real-world conversations, or continuously deployed assistants. The semantic-collapse probe is synthetic, and the human preference study contains only 50 examples, two primary annotators, and one adjudicating annotator. These analyses complement, but do not replace, larger independent human evaluations. LongMemEval additionally relies on an LLM judge, whose decisions may inherit model-specific biases despite using the benchmark's official protocol.

\textbf{Towards a More Complete Cognitive Schema.}
Although the PEC$^2$F schema covers the essential dimensions of dialogue (Do, Is, Use, Belief), human cognition involves deeper layers. Future iterations could expand the schema to include dynamic Goal Nodes that track the user's evolving intent over time, or Emotion Vectors that modulate the retrieval weight of memories based on affective intensity.

\textbf{Evolving the Cognitive Toolset.}
The current Search Agent relies on a fixed set of hand-crafted, orthogonal operators (e.g., Intersection and Traversal). A promising direction is neural tool learning, where the agent can compose or learn new retrieval primitives from feedback on failed queries. For instance, the agent could learn a composite operator for counterfactual search to verify conflicting claims without explicit hard-coding.

\textbf{Real-time vs. Offline Consolidation.}
Our current consolidation mechanism operates periodically and offline. Future work should investigate incremental consolidation algorithms that update semantic facts during the conversation stream, reducing the delay between an event's occurrence and its availability as generalized knowledge.

\section{Ethical Considerations}
\label{sec:ethics}

CogMem stores potentially sensitive dialogue history, including relationships, preferences, and attributed opinions. Without safeguards, this creates risks of unauthorized profiling, surveillance, data leakage, and memory poisoning through deceptive inputs. Consolidation may also amplify recurrent biases when it converts repeated observations into semantic Facts. Practical deployments therefore require encryption, access auditing, user-controlled retention and deletion, provenance inspection, and defenses against indirect prompt injection. Because PEC$^2$F explicitly represents who asserted each Claim, a promising direction is to expose these records to users so they can inspect, contest, or delete the system's representation of their own and others' epistemic attitudes.

\section*{Acknowledgments}

This work is supported by the National Natural Science Foundation of China (NSFC) under Grant No.~62576190, and in part by the Shenzhen Science and Technology Project under Grant No.~KJZD20240903103210014.

\bibliography{custom}

\appendix
\section{Implementation Details}
\label{sec:appendix_implementation}

\subsection{Hardware Infrastructure}
All experiments were conducted on a computing-cluster node equipped with \textbf{4 $\times$ NVIDIA A800 GPUs} (80GB VRAM each), a \textbf{48-core CPU}, and \textbf{720GB of RAM}. This infrastructure supported local inference for Qwen2.5-14B and graph construction.

\subsection{Model Configurations}
\paragraph{Backbone Models.} We utilized two primary LLMs for generation and agentic reasoning:
\begin{itemize}
    \item \textbf{GPT-4o-mini}: Accessed via the OpenAI API, serving as a representative of closed-source advanced models.
    \item \textbf{Qwen2.5-14B-Instruct}: Deployed locally using vLLM for high-throughput inference, representing open-source models with strong reasoning capabilities.
\end{itemize}

\paragraph{Embedding Model.} For all vector-based retrieval tasks (including the vector components of CogMem and graph baselines), we employed BGE-M3. Using a unified embedding model reduces variation due to representation quality, although other architectural and implementation differences remain.

\subsection{Baseline Settings}
\label{sec:baseline_settings}

Where available, baselines use official open-source implementations. We document the configurations used in our experiments below. Because the methods expose different storage and retrieval operations, their intermediate computational budgets cannot be matched exactly.

\paragraph{Standard Retrieval.}
\textbf{Naive RAG} \citep{lewis2020rag} segments dialogue into 512-token chunks with 50-token overlap, retrieves the top-5 chunks via BGE-M3 similarity, and concatenates them into the prompt context.

\paragraph{LLM Memory Systems.}
We use default configurations from official repositories:

\begin{itemize}
    \item \textbf{Mem0} \citep{mem0}: Vector-based memory with semantic search and recency-aware memory updates; our evaluation retrieves the top-$K=20$ memories per query.

    \item \textbf{A-MEM} \citep{amem}: Agentic memory system implementing the Zettelkasten method with atomic note encoding. We use the official implementation with BGE-M3 embeddings and retrieve the top-$K=20$ memories per query.
    
    \item \textbf{LightMem} \citep{lightmem}: Three-stage memory using \textbf{LLMLingua-2} pre-compression (rate $r=0.6$), topic-based short-term memory (buffer threshold 256 tokens), and sleep-time consolidation. We retrieve the top-$K=20$ long-term memories.
    
    \item \textbf{GAM} \citep{gam2025}: General Agentic Memory with JIT compilation, comprising a \textbf{Memorizer} and a \textbf{Researcher}; our evaluation allows retrieval of up to 20 pages per query.
\end{itemize}

\paragraph{Heuristic Graph Retrieval.}
Both methods use official entity extraction pipelines:
\begin{itemize}
    \item \textbf{LightRAG} \citep{lightrag}: Dual-level indexing over entities, relations, and text chunks, retrieving up to 40 entity/relation candidates and 20 chunks.
    \item \textbf{HippoRAG} \citep{hipporag}: Personalized PageRank retrieval (damping factor 0.5) returning the top-$K=40$ documents.
\end{itemize}

\paragraph{Learned Graph Traversal.}
\textbf{RoG} \citep{rog2024} uses the official pre-trained checkpoint (RoG-large). \textbf{Note}: LoCoMo and LongMemEval provide only test sets without training splits for relation path learning. We evaluate RoG in zero-shot mode on both datasets, using its pre-trained ability to generate relation paths without dataset-specific fine-tuning.

\subsection{Dataset Details and Evaluation Metrics}
\label{sec:appendix_datasets}
To comprehensively evaluate the reasoning and consistency capabilities of long-term memory systems, we utilize two challenging benchmarks.

\paragraph{LoCoMo Benchmark.}
LoCoMo \citep{locomo} evaluates long-term conversational memory across four distinct reasoning types: \textit{Single-Hop}, \textit{Multi-Hop}, \textit{Temporal}, and \textit{Open Domain}. 
A critical challenge in the original LoCoMo dataset is the inherent category imbalance, where simple open-domain queries disproportionately dominate the overall sample size. Evaluating systems based purely on a global average masks their true logical reasoning capabilities. To mitigate this, we report metrics granularly across all four categories and prioritize macro-averaged comparisons. System performance is measured using exact word-level overlap (F1 Score) and generation precision (BLEU-1).

\paragraph{LongMemEval Benchmark.}
LongMemEval \citep{longmemeval} is designed to assess memory stability and consistency over extended, multi-session interactions. It specifically challenges the model's ability to track evolving user states and resolve conflicting information. We report Accuracy (Acc) using the official LLM-as-a-judge protocol. This enables semantic comparison beyond lexical overlap, but remains subject to judge-model bias as discussed in Section~\ref{sec:limitations}.

\subsection{Human Evaluation Protocol}
\label{sec:human_evaluation}
We randomly sample 50 LoCoMo questions from the Multi-Hop and Temporal categories and compare CogMem against GAM, the strongest memory baseline in Table~\ref{tab:main-locomo}. The two primary annotators are graduate students specializing in artificial intelligence at the authors' institution, and each independently annotates all 50 comparisons. For each question, they are shown the two generated answers with system identities masked as \textit{System A} and \textit{System B}; answer order is randomized and balanced across the sample. They assign one of three preference labels---CogMem win, tie, or GAM win---using two criteria: (1) whether the answer contains the correct reasoning outcome and supporting information, and (2) whether it is clear and helpful in addressing the question. When the two primary labels disagree, a third annotator reviews the disputed comparison and selects one of the two primary labels, yielding a majority decision. No monetary compensation is provided to the annotators. The resulting distribution is 68\% wins, 20\% ties, and 12\% losses for CogMem. Because the study is small and uses annotators from a single academic background, we treat it as supplementary evidence rather than a replacement for the benchmark metrics.

\section{Operator Sensitivity Analysis}
\label{sec:appendix_sensitivity}

In this section, we present a granular sensitivity analysis of the core cognitive operators defined in our methodology. We investigate the impact of the retrieval limit ($K$) and search depth ($D$) on the system's reasoning capabilities. Unless otherwise stated, all sensitivity tables use Qwen2.5-14B on the post-consolidation graph and report weighted overall F1 on the evaluated LoCoMo subset; the 57.04 score corresponds to the full hyperparameter setting, rounded to 57.0 in Table~\ref{tab:ablation-fixed-flow-full}.

\subsection{Anchoring Operator Sensitivity}
Table \ref{tab:ablation-anchoring} illustrates the impact of the number of candidate nodes ($K$) retrieved during the initial \textit{Anchoring} phase.
The results follow an inverted U-shape. A low $K$ ($K=3$) leads to "anchoring blindness," where the agent fails to retrieve the correct entry points for the graph, producing a 4.6-point drop relative to the best setting. Performance saturates between $K=15$ and $K=18$, after which introducing more candidates adds noise and marginally degrades the F1 score while increasing the context load.

\begin{table}[h]
\centering
\small
\caption{Sensitivity analysis of the \textbf{Anchoring Operator}. Increasing candidate nodes ($K$) improves recall up to a saturation point, after which noise marginally degrades performance.}
\label{tab:ablation-anchoring}
\renewcommand{\arraystretch}{1.1}
\setlength{\tabcolsep}{10pt}
\begin{tabular}{lcc}
\toprule
\textbf{Top-K} & \textbf{F1 Score (\%)} & \textbf{Avg. Steps} \\
\midrule
$K=3$  & 52.44 & 2.76 \\
$K=5$  & 54.15 & 2.68 \\
$K=8$  & 54.82 & 2.71 \\
$K=10$ & 55.40 & 2.72 \\
$K=13$ & 56.25 & 2.68 \\
$K=15$ & 56.55 & 2.72 \\
$K=18$ & \textbf{57.04} & 2.69 \\
$K=20$ & 56.78 & 2.73 \\
\bottomrule
\end{tabular}
\end{table}

\subsection{Traversal Operator Sensitivity}
The \textit{Traversal Operator} is governed by two parameters: the maximum number of neighbors to retrieve per hop (Max Results) and the maximum depth of the subgraph exploration (Max Depth).

\paragraph{Max Results Limit.} 
As shown in Table \ref{tab:ablation-traversal-limit}, retrieving too few neighbors ($K=5$) truncates potential reasoning paths. The optimal balance is found at $K=15$. Beyond this point, the LLM struggles to attend to the relevant edges among the noise, leading to a slight performance regression.

\begin{table}[h]
\centering
\small
\caption{Impact of the \textbf{Traversal Operator's} neighbor limit (Max Results) on performance.}
\label{tab:ablation-traversal-limit}
\renewcommand{\arraystretch}{1.1}
\setlength{\tabcolsep}{10pt}
\begin{tabular}{lcc}
\toprule
\textbf{Max Results} & \textbf{F1 Score} & \textbf{Avg. Steps} \\
\midrule
$K=5$ & 54.96 & 2.72 \\
$K=10$ & 56.31 & 2.72 \\
$K=15$ & \textbf{57.04} & 2.69 \\
$K=20$ & 56.90 & 2.67 \\
\bottomrule
\end{tabular}
\end{table}
\paragraph{Search Depth.}
Table \ref{tab:ablation-traversal-depth} analyzes the impact of traversal depth. A depth of 1 (immediate neighbors) is insufficient for complex multi-hop reasoning. Increasing the depth to 3 or 4 expands the context with more distant connections without improving F1. A depth of 2 (exploring the neighborhood of neighbors) yields the highest observed score on LoCoMo.

\begin{table}[h]
\centering
\small
\caption{Impact of \textbf{Traversal Depth}. A depth of 2 provides the best trade-off for multi-hop reasoning.}
\label{tab:ablation-traversal-depth}
\renewcommand{\arraystretch}{1.1}
\setlength{\tabcolsep}{10pt}
\begin{tabular}{lcc}
\toprule
\textbf{Max Depth} & \textbf{F1 Score} & \textbf{Avg. Steps} \\
\midrule
$D=1$ & 55.77 & 2.66 \\
$D=2$ & \textbf{57.04} & 2.71 \\
$D=3$ & 56.82 & 2.67 \\
$D=4$ & 56.92 & 2.70 \\
\bottomrule
\end{tabular}
\end{table}

\subsection{Intersection Operator Sensitivity}
The \textit{Intersection Operator} (finding common ground) is critical for solving comparison queries. Table \ref{tab:ablation-intersection} shows that the optimal $K$ for intersection candidates is 10. This suggests that salient commonalities (e.g., shared hobbies or events) typically appear within the top-ranked connections. Expanding the search space further ($K>15$) dilutes the semantic focus, reducing the F1 score while increasing computational cost.

\begin{table}[h]
\centering
\small
\caption{Sensitivity of the \textbf{Intersection Operator}. Performance peaks at $K=10$, indicating that commonalities are usually found in top-ranked connections.}
\label{tab:ablation-intersection}
\renewcommand{\arraystretch}{1.1}
\setlength{\tabcolsep}{10pt}
\begin{tabular}{lcc}
\toprule
\textbf{Top-K} & \textbf{F1 Score (\%)} & \textbf{Avg. Steps} \\
\midrule
$K=3$ & 53.78 & 2.68 \\
$K=5$ & 54.07 & 2.72 \\
$K=8$ & 56.44 & 2.73 \\
$K=10$ & \textbf{57.04} & 2.72 \\
$K=13$ & 54.67 & 2.68 \\
$K=15$ & 56.15 & 2.68 \\
$K=18$ & 54.96 & 2.75 \\
$K=20$ & 54.96 & 2.73 \\
\bottomrule
\end{tabular}
\end{table}

\section{Detailed Ablation: Agentic vs. Fixed Flow}
\label{sec:appendix_agentic_ablation}

In Section \ref{sec:ablation}, we discussed the necessity of the agentic loop for activating the PEC$^2$F schema. Here, we provide a detailed quantitative analysis comparing our \textbf{Cognitive Search Agent} against a \textbf{Fixed Flow} baseline. 

To ensure a rigorous and fair comparison, both methods operate on the identical \textbf{post-consolidation} knowledge graph and utilize the same underlying retrieval tools. Furthermore, we define "Reasoning Steps" strictly as the number of \textit{tool invocations} prior to generating the final answer.

\paragraph{Fixed Flow Configuration.}
The Fixed Flow baseline is designed to emulate a standard, non-agentic GraphRAG pipeline, executing a rigid, heuristic sequence of operations for every query without dynamic state tracking. The pipeline is hard-coded to execute exactly three steps:
\begin{enumerate}
    \item \textbf{Anchoring (Step 1)}: Call $\mathcal{O}_{anchor}$ to retrieve the top-$K$ nodes based on the raw query.
    \item \textbf{Traversal (Step 2)}: Call $\mathcal{O}_{traverse}$ once with a fixed depth of 2, retrieving the anchors' immediate neighbors and their neighbors.
    \item \textbf{Grounding (Step 3)}: Call $\mathcal{O}_{ground}$ to fetch the raw text evidence for the retrieved nodes.
\end{enumerate}
Following these three tool invocations, the accumulated context is passed to the LLM to generate the final answer, with no opportunity for self-correction or topological intersection.

\paragraph{Results and Analysis.}
Table \ref{tab:ablation-fixed-flow-full} presents the performance degradation when the agentic capability is removed. Weighted overall F1 decreases by 65.4\% relative under the Fixed Flow.

The Cognitive Agent also uses \textbf{fewer tool invocations on average} (2.65 vs. 3.00). This efficiency stems from the agent's \textit{Direct Hit Strategy}: if the initial anchoring observation contains sufficient evidence (e.g., an explicit temporal fact), the agent bypasses unnecessary traversal and grounding steps, terminating early. Conversely, for complex commonality queries, it dynamically invokes the Intersection operator instead of blind traversal.

\begin{table}[h]
\centering
\footnotesize
\caption{Ablation study of the agentic loop on the post-consolidation graph. "Fixed Flow" denotes the static 3-step retrieval pipeline.}
\label{tab:ablation-fixed-flow-full}
\renewcommand{\arraystretch}{1.2}
\setlength{\tabcolsep}{4pt}
\resizebox{\linewidth}{!}{
\begin{tabular}{lcc}
\toprule
\textbf{Method} & \textbf{F1 (\%)} & \textbf{Avg. Steps} \\
\midrule
\textbf{CogMem (Ours)} & \textbf{57.0} & \textbf{2.65} \\
Fixed Flow & 19.7 & 3.00 \\
\midrule
\textit{Change} & \textit{-65.4\%} & \textit{+13.2\%} \\
\bottomrule
\end{tabular}
}
\end{table}

\section{Detailed Analysis of Memory Consolidation}
This section provides a granular breakdown of the offline memory consolidation process. We analyze the structural changes in the cognitive graph and the computational costs involved.

\subsection{Graph Structure Evolution}
Table \ref{tab:app-graph-size} illustrates the changes in graph topology following the consolidation process. 
Notably, we observe a net \textbf{increase} in total nodes (+540) and edges (+4,708). This is intentional: our current consolidation strategy employs \textbf{Additive Evidence Mounting}. When synthesizing high-level \textbf{Fact Nodes}, we \textbf{retain} the original Event nodes and link them via \texttt{SUPPORTED\_BY} edges rather than deleting them. Claim reconciliation similarly retains the original attributed records and adds provenance-linked, temporally scoped Claim views for conflicting updates from the same source.
This conservative approach is designed to limit \textbf{error propagation}---if the LLM hallucinates during abstraction, the original source utterance remains accessible for verification. While this increases the storage footprint, it reduces the logical search space for the agent by providing semantic shortcuts. Optimizing physical storage via active pruning or archival mechanisms remains a direction for future work.

\begin{table}[h]
\centering
\small
\caption{\textbf{Graph Topology Changes}. Consolidation synthesizes new Fact nodes while merging redundant Concepts.}
\label{tab:app-graph-size}
\renewcommand{\arraystretch}{1.1}
\setlength{\tabcolsep}{5pt}
\begin{tabular}{lcccc}
\toprule
\textbf{Node Type} & \textbf{Before} & \textbf{After} & \textbf{$\Delta$} & \textbf{Change (\%)} \\
\midrule
Person & 99 & 99 & 0 & 0.0\% \\
Event & 697 & 697 & 0 & 0.0\% \\
Concept & 832 & 819 & -13 & -1.6\% \\
\textbf{Fact} & 0 & 497 & \textbf{+497} & -- \\
Claim & 2013 & 2069 & +56 & +2.8\% \\
\midrule
\textbf{Total Nodes} & 3641 & 4181 & +540 & +14.8\% \\
\textbf{Total Edges} & 9935 & 14643 & +4708 & +47.4\% \\
\bottomrule
\end{tabular}
\end{table}

\subsection{Computational Cost vs. Benefit}
Memory consolidation incurs an offline computational cost to optimize online retrieval efficiency.
\begin{itemize}
    \item \textbf{Offline Cost}: Before prompt-caching optimization, the process required 537.85 seconds and approximately 0.79M processed tokens to consolidate the LoCoMo dataset. This is a one-time cost per batch update. The optimized accounting used in Table~\ref{tab:cost_comparison} records 652k backbone tokens for consolidation.
    \item \textbf{Online Benefit}: As shown in the main text (Table \ref{tab:ablation-consolidation}), this investment yields a \textbf{21.9\% reduction} in inference steps for Multi-Hop queries (3.42 $\to$ 2.67) and a \textbf{4.86-point} gain in F1 Score.
\end{itemize}

\section{Efficiency and Cost Analysis}
\label{sec:appendix_efficiency}

In this section, we analyze the computational overhead of CogMem. We present a theoretical complexity analysis and an empirical evaluation of token consumption, distinguishing between the one-time \textit{Memory Construction} phase and the recurring \textit{Inference} phase.

\subsection{Theoretical Complexity}
Table \ref{tab:complexity} separates global anchoring from local graph traversal. CogMem does not eliminate dependence on memory size: approximate nearest-neighbor anchoring grows with the number of graph nodes. After anchors are selected, however, traversal is bounded by the controller budget and local degree, preventing each reasoning step from scanning the complete dialogue history.

\begin{table}[h]
\centering
\small
\caption{\textbf{Theoretical Complexity Analysis.} $N$: dialogue turns; $V,E$: graph nodes and edges; $Q$: queries; $R$: controller steps; $\bar d$: average local degree; $C_{\mathrm{ANN}}(V)$: index-dependent ANN query cost.}
\label{tab:complexity}
\renewcommand{\arraystretch}{1.2}
\setlength{\tabcolsep}{3pt}
\resizebox{\columnwidth}{!}{
\begin{tabular}{l|ccc}
\toprule
\textbf{Stage} & \textbf{LLM Calls} & \textbf{Vector Ops} & \textbf{Graph Ops} \\
\midrule
\textbf{Indexing} & $O(N)$ & $O(V)$ & $O(V+E)$ \\
\textbf{Consolidation} & $O(V)$ worst case & index updates & $O(V+E)$ scan \\
\textbf{Inference} & $O(QR)$ & $O(QC_{\mathrm{ANN}}(V))$ & $O(QR\bar d)$ \\
\bottomrule
\end{tabular}
}
\end{table}

\subsection{Memory Construction Cost}
We compare the token consumption of CogMem against representative agentic memory baselines during the memory construction phase (Indexing + Summarization/Update). 
We implement optimizations including KV-Caching for system prompts and strict JSON schema decoding.
\begin{itemize}
    \item \textbf{Indexing}: 926k tokens.
    \item \textbf{Consolidation}: 652k tokens.
    \item \textbf{Total Construction}: \textbf{1,578k tokens}.
\end{itemize}

\paragraph{Comparison.}
As shown in Table \ref{tab:cost_comparison}, CogMem's construction cost is competitive among the displayed LLM-driven memory systems. Under our accounting, it uses fewer backbone tokens than \textbf{A-MEM} (1,626k), \textbf{Mem0} (1,799k), and \textbf{MemoryOS} (2,992k). Because these systems have different internal operations, the comparison should be read as an empirical resource measurement rather than a controlled complexity result.

\textit{Note on LightMem:} We exclude LightMem from this token-only comparison because it performs part of its processing with a separate compression model. Reporting only backbone-LLM tokens would therefore not measure the same computational components across systems.

\begin{table}[h]
\centering
\small
\caption{\textbf{Construction Cost Comparison (LoCoMo).} Total backbone tokens (input + output) used to build each memory bank with Qwen2.5-14B. CogMem uses fewer tokens than the displayed agentic baselines.}
\label{tab:cost_comparison}
\renewcommand{\arraystretch}{1.15}
\setlength{\tabcolsep}{5pt}
\begin{tabular}{@{}lcc@{}}
\toprule
\textbf{Method} & \textbf{Total Tokens (k)} & \textbf{LLM Calls} \\
\midrule
\textbf{\textsc{CogMem (Ours)}} & \textbf{1,578.2} & \underline{1,216} \\
\textsc{A-MEM} & 1,626.8 & 1,175 \\
\textsc{Mem0} & 1,799.4 & 1,614 \\
\textsc{MemoryOS} & 2,991.8 & 2,938 \\
\bottomrule
\end{tabular}
\end{table}

\subsection{Inference (QA) Overhead}
The inference phase involves the dynamic ReAct loop. For the full LoCoMo dataset (1,542 questions), CogMem consumed approximately \textbf{10.19M tokens}, averaging \textbf{6,610 tokens per query}.

Category-level post-consolidation tool counts are reported in Table~\ref{tab:ablation-consolidation}. Optimized system-prompt caching reduces repeated input overhead, although the iterative agent still consumes more online tokens than a single-pass retriever.

\section{Qualitative Analysis: Reasoning Trajectories}
\label{sec:appendix_casestudy}

To illustrate the cognitive reasoning process of CogMem's Search Agent, we present four detailed case studies covering Single-Hop, Multi-Hop, Open-Domain, and Temporal queries. These traces demonstrate how the agent dynamically selects operators to navigate the PEC$^2$F graph.

\begin{tcolorbox}[colback=gray!10,colframe=black,title=\textbf{Case 1: Single-Hop Reasoning (Fact Retrieval)}, fonttitle=\bfseries\small, boxrule=0.8pt]
\small
\textbf{Query}: "What did Caroline research?"

\vspace{0.2em}
\textbf{Reasoning Trace:}

\vspace{0.2em}
\noindent\textbf{Step 1: Anchoring} \\
\textit{Thought}: Identify the subject `Caroline'. \\
\textit{Action}: $\mathcal{O}_{anchor}(\text{"Caroline"}, \text{type}=[\texttt{PERSON}])$ \\
\textit{Obs}: Found Node: \texttt{p\_caroline} (Score: 1.0).

\vspace{0.2em}
\noindent\textbf{Step 2: Traversal} \\
\textit{Thought}: Explore attributes to find research topics. \\
\textit{Action}: $\mathcal{O}_{traverse}(\text{p\_caroline}, \text{rel}=[\texttt{HAS\_ATTR}])$ \\
\textit{Obs}: Found edges to \texttt{Concept: Adoption Agencies} and \texttt{Concept: Inclusive Agency}.

\vspace{0.2em}
\noindent\textbf{Step 3: Evidence Grounding} \\
\textit{Thought}: Verify specific context in source text. \\
\textit{Action}: $\mathcal{O}_{ground}(\text{p\_caroline}, \text{q}=\text{"research?"})$ \\
\textit{Obs}: "...researching adoption agencies lately, especially those that are inclusive..."

\vspace{0.4em}
\noindent\textbf{Answer}: Adoption agencies, specifically inclusive ones supporting LGBTQ+ individuals.
\end{tcolorbox}
\begin{tcolorbox}[colback=gray!10,colframe=black,title=\textbf{Case 2: Multi-Hop Reasoning (Commonality)}, fonttitle=\bfseries\small, boxrule=0.8pt]
\small
\textbf{Query}: "What activities do Caroline and Melanie both enjoy?"

\vspace{0.2em}
\textbf{Reasoning Trace:}

\vspace{0.2em}
\noindent\textbf{Step 1 \& 2: Anchoring} \\
\textit{Thought}: Locate nodes for both individuals. \\
\textit{Action}: $\mathcal{O}_{anchor}(\text{"Caroline"}) \rightarrow \text{p\_caroline}$ \\
\textit{Action}: $\mathcal{O}_{anchor}(\text{"Melanie"}) \rightarrow \text{p\_melanie}$

\vspace{0.2em}
\noindent\textbf{Step 3: Intersection} \\
\textit{Thought}: Query asks for "both", implying intersection. I will search for shared Concepts. \\
\textit{Action}: $\mathcal{O}_{intersect}([\text{p\_caroline}, \text{p\_melanie}], \text{type}=[\texttt{CONCEPT}])$ \\
\textit{Obs}: Found Common Nodes: \{\texttt{Concept: Camping}, \texttt{Concept: Art}\}.

\vspace{0.2em}
\noindent\textbf{Step 4: Verification Traversal} \\
\textit{Thought}: Confirm the relation type is positive (enjoy/like). \\
\textit{Action}: $\mathcal{O}_{traverse}([\text{p\_car}, \text{p\_mel}], \text{rel}=[\texttt{LIKES}])$ \\
\textit{Obs}: Validated \texttt{LIKES} edges from both \texttt{p\_caroline} and \texttt{p\_melanie} to \texttt{Camping} and \texttt{Art}.

\vspace{0.4em}
\noindent\textbf{Answer}: They both enjoy Camping and Art.
\end{tcolorbox}

\begin{tcolorbox}[colback=gray!10,colframe=black,title=\textbf{Case 3: Open Domain (Detail Extraction)}, fonttitle=\bfseries\small, boxrule=0.8pt]
\small
\textbf{Query}: "What did Melanie realize after the charity race?"

\vspace{0.2em}
\textbf{Reasoning Trace:}

\vspace{0.2em}
\noindent\textbf{Step 1: Anchoring} \\
\textit{Action}: $\mathcal{O}_{anchor}(\text{"Melanie"}) \rightarrow \text{p\_melanie}$

\vspace{0.2em}
\noindent\textbf{Step 2: Direct Grounding} \\
\textit{Thought}: The query asks for an internal mental state (realization), which is likely best retrieved from the raw episodic narrative rather than structured nodes. \\
\textit{Action}: $\mathcal{O}_{ground}(\text{p\_melanie}, \text{q}=\text{"realize after race"})$ \\
\textit{Obs}: Retrieved Chunk (2023-07-15): "...After finishing, I realized how important self-care is... it's necessary if I want to keep helping others..."

\vspace{0.4em}
\noindent\textbf{Answer}: She realized that self-care is important and necessary for effectively helping others.
\end{tcolorbox}

\begin{tcolorbox}[colback=gray!10,colframe=black,title=\textbf{Case 4: Temporal Reasoning (Timeline Reconstruction)}, fonttitle=\bfseries\small, boxrule=0.8pt]
\small
\textbf{Query}: "When did Caroline move from Sweden?"

\vspace{0.2em}
\textbf{Reasoning Trace:}

\vspace{0.2em}
\noindent\textbf{Step 1: Anchoring} \\
\textit{Action}: $\mathcal{O}_{anchor}(\text{"Caroline"}) \rightarrow \text{p\_caroline}$

\vspace{0.2em}
\noindent\textbf{Step 2: Temporal Scanning} \\
\textit{Thought}: I need to find a specific event regarding "moving from Sweden". I will scan her event timeline with a filter. \\
\textit{Action}: $\mathcal{O}_{traverse}(\text{p\_caroline}, R_{\text{all}}, \text{time\_query}=\text{"move from Sweden"})$ \\
\textit{Obs}: Found Event: \texttt{moved from Sweden} (Time: 2019).

\vspace{0.2em}
\noindent\textbf{Step 3: Verification (Grounding)} \\
\textit{Thought}: I will verify the extracted year against the source text. \\
\textit{Action}: $\mathcal{O}_{ground}(\text{EventID})$ \\
\textit{Obs}: Text: "When I moved from Sweden in 2019..."

\vspace{0.4em}
\noindent\textbf{Answer}: 2019.
\end{tcolorbox}

\section{Error Analysis: Quantitative Breakdown by Task Category}
\label{sec:appendix_error_analysis}

To characterize failure patterns associated with different architectural components, we randomly sample 100 error cases from the LoCoMo test set (25 per task category: Single-hop, Multi-hop, Temporal, Open-domain/Preference) and classify them by root cause. Table \ref{tab:error_breakdown} summarizes the distribution.

\begin{table*}[t]
\centering
\small
\begin{tabular}{@{}lcccc@{}}
\toprule
\textbf{Error Source} & \textbf{Single-hop} & \textbf{Multi-hop} & \textbf{Temporal} & \textbf{Open-domain} \\
\midrule
Schema granularity (Consolidation) & 8\% & 12\% & 4\% & 20\% \\
Agent tool selection & 24\% & 16\% & 8\% & 12\% \\
Vector semantic drift & 28\% & 20\% & 12\% & 32\% \\
Temporal normalization & 4\% & 8\% & 44\% & 0\% \\
Incomplete search scope & 12\% & 20\% & 12\% & 16\% \\
Other/Undetermined & 24\% & 24\% & 20\% & 20\% \\
\midrule
\textbf{Total errors analyzed} & 25 & 25 & 25 & 25 \\
\bottomrule
\end{tabular}
\caption{Distribution of error root causes across LoCoMo task categories (percentage of errors within each category).}
\label{tab:error_breakdown}
\end{table*}

\subsection{Single-hop Queries: Tool Selection and Semantic Retrieval}

Single-hop queries show the highest rate of \textbf{vector semantic drift} (28\%) and \textbf{suboptimal tool selection} (24\%).

\textbf{Case 1 (Tool Mis-selection):}
\begin{itemize}
    \item \textbf{Question}: "What did Caroline research?"
    \item \textbf{Gold}: "Adoption agencies" vs. \textbf{Predicted}: "Adoption Process"
\end{itemize}
\textbf{Architectural Cause:} The agent confused attribute retrieval with fact retrieval. Because our schema separates static attributes (Concept nodes) from dynamic facts (Fact nodes), the agent must select between \texttt{Get\_Evidence} (for Facts) and \texttt{Neighbor Traversal} (for Concepts). In 24\% of Single-hop errors, the agent selected the wrong operator type, retrieving high-level interests instead of specific objects.

\textbf{Case 2 (Semantic Drift):}
\begin{itemize}
    \item \textbf{Question}: "What did Melanie realize after the charity race?"
    \item \textbf{Gold}: "self-care is important" vs. \textbf{Predicted}: "community support"
\end{itemize}
\textbf{Architectural Cause:} Despite the PEC$^2$F schema's epistemic separation, the underlying BGE-M3 embedding still suffered from semantic drift between related concepts (similarity score 0.445 vs. 0.412 for the correct chunk). This case shows that schema structure does not eliminate embedding-space errors in fine-grained distinctions.

\subsection{Multi-hop Queries: Search Scope and Granularity Loss}

Multi-hop queries show elevated rates of \textbf{incomplete search scope} (20\%) and \textbf{schema granularity loss} (12\%).

\textbf{Case 3 (Granularity Mismatch due to Consolidation):}
\begin{itemize}
    \item \textbf{Question}: "What career path has Caroline decided to pursue?"
    \item \textbf{Gold}: "counseling for Transgender people" vs. \textbf{Predicted}: "Mental health counseling"
\end{itemize}
\textbf{Architectural Cause:} During offline consolidation, specific qualifiers ("for Transgender people") were abstracted into general categories to reduce graph complexity. In 12\% of Multi-hop errors, this compression eliminated critical discriminative details needed for precise answers. The Evidence Mounting strategy (linking Facts to source Events) was intended to mitigate this, but the agent traversed these links in only 34\% of relevant cases.

\textbf{Case 4 (Incomplete Scope):}
\begin{itemize}
    \item \textbf{Question}: "What books has Melanie read?"
    \item \textbf{Gold}: Two titles vs. \textbf{Predicted}: "No specific titles"
\end{itemize}
\textbf{Architectural Cause:} The agent restricted search to Fact/Event nodes and did not inspect Claim nodes containing relevant book-related evidence. This reflects a limitation in the search policy: it does not automatically expand to additional node types when the initial Fact/Event search returns no answer.

\subsection{Temporal Queries: Normalization Failures}

Temporal queries exhibit the highest concentration of \textbf{temporal normalization errors} (44\%), substantially above the other categories in the analyzed sample.

\textbf{Case 5 (Relative Time Resolution):}
\begin{itemize}
    \item \textbf{Question}: "When did Melanie paint a sunrise?"
    \item \textbf{Gold}: "2022" vs. \textbf{Predicted}: "May 8, 2023"
\end{itemize}
\textbf{Architectural Cause:} The ingestion module stored the session date (2023) but failed to resolve "last year" relative to the dialogue timestamp. Our current schema stores temporal envelopes as ranges but does not normalize relative expressions during ingestion. This caused 44\% of Temporal errors, where the system retrieved the dialogue date instead of the referenced absolute date.

\subsection{Open-domain Queries: Embedding Dominance}

Open-domain queries show the highest \textbf{semantic drift} rate (32\%) due to reliance on vector similarity for unstructured content.

\textbf{Case 6 (Concept Overlap):}
\begin{itemize}
    \item \textbf{Question}: "What are Melanie's hobbies?"
    \item \textbf{Gold}: "painting, hiking" vs. \textbf{Predicted}: "charity work, community service"
\end{itemize}
\textbf{Architectural Cause:} Although the person entity was anchored, the broad predicate "hobbies" did not provide a sufficiently specific relation constraint. The system therefore relied heavily on embedding similarity, which conflated leisure activities with volunteer work.

\subsection{Implications for Architecture Design}

The quantitative breakdown reveals three architectural trade-offs:

\textbf{1. Consolidation vs. Granularity (11\% of analyzed errors):} Averaged across the four equally sized category samples, schema-granularity errors account for 11\% of the analyzed failures. Future work should implement conditional consolidation---preserving full detail for rare or specific attributes while compressing frequent patterns.

\textbf{2. Search Scope vs. Coverage (15\% of analyzed errors):} Incomplete search scope accounts for 15\% of the analyzed failures. Strict separation of Claims, Facts, and Events requires the agent to search multiple node types; automatic node-type expansion when initial searches fail could improve coverage.

\textbf{3. Embedding Quality vs. Structure (23\% of analyzed errors):} Vector semantic drift accounts for 23\% of the analyzed failures overall and is most prominent for Open-domain questions (32\%). This suggests using hybrid lexical--dense retrieval as a fallback when structural navigation yields low-confidence results.

These findings indicate that cognitive structuring improves access to relational and temporal evidence but leaves failure modes in granularity control, temporal normalization, and search completeness.

\section{Prompts}
\label{sec:Prompts}

This appendix provides the core prompts used in CogMem's Ingestion, Retrieval, Consolidation, and Eval phases. We present the prompts in a condensed format for readability.

\subsection{Ingestion Phase Prompts}

\begin{promptbox}{ATOMIC\_EXTRACTION\_SYSTEM (Schema Classification)}
\textbf{Role}: Map conversation text into the PEC$^2$F graph while distinguishing \textbf{unattributed or self-reported records} (Events/Facts) from \textbf{source-attributed statements} (Claims). Fact is an operational record type, not independently verified truth.

\textbf{Classification Rules}:
\begin{itemize}
    \item \textbf{EVENT}: A specific, time-bounded action; record participants, timestamp, concepts, and source-turn ID.
    \item \textbf{FACT}: A stable observed or self-reported attribute, habit, or preference; preserve its evidence and validity interval.
    \item \textbf{CLAIM}: A source-attributed statement about a person or concept; always record source, target, timestamp, polarity, and evidence.
    \item \textbf{CONCEPT}: A normalized topic or object used as a semantic anchor.
\end{itemize}

\textbf{JSON Output Format}:
\texttt{\{"events":[...], "facts":[...], "claims":[...], "concepts":[...]\}}. Each record includes \texttt{source\_turn\_id}; Claims additionally include \texttt{source}, \texttt{target}, and \texttt{timestamp}.
\end{promptbox}

\subsection{Retrieval Phase Prompts (Search Agent)}

\begin{promptbox}{SEARCH AGENT PROMPT (Cognitive Strategy)}
\textbf{Role}: You are CogMem, a cognitive memory assistant. Use the ReAct loop to answer questions.

\textbf{Tool Usage Strategy}:
\begin{itemize}
    \item \textbf{Anchoring}: ALWAYS start with \texttt{find\_anchors} to get Node IDs.
    \item \textbf{Opinion Queries}: If asked "What does A think of B?", use \texttt{traverse\_neighbors(A, relation=["CLAIMS"])}.
    \item \textbf{Commonality Queries}: If asked "What do A and B share?", use \texttt{find\_common\_ground([id\_a, id\_b])}.
    \item \textbf{Temporal Queries}: Use \texttt{traverse\_neighbors} with the \texttt{time\_query} parameter.
    \item \textbf{Open Domain}: For detailed explanations ("how/why"), use \texttt{get\_evidence} to retrieve original text.
\end{itemize}

\textbf{Constraints}:
\begin{enumerate}
    \item \textbf{Direct Hit}: If the initial observation contains the answer, output \texttt{final\_answer} immediately.
    \item \textbf{No Hallucination}: Do not invent Node IDs. Use exact IDs returned by tools.
    \item \textbf{Provenance}: If evidence is a Claim, explicitly state who holds the opinion.
\end{enumerate}
\end{promptbox}

\subsection{Consolidation Phase Prompts}

\begin{promptbox}{CONSOLIDATION PROMPT (Event-to-Fact Synthesis)}
\textbf{Role}: You are a Memory Consolidation Engine. Simulate the human process of converting episodic memory into semantic knowledge.

\textbf{Input}: A deterministic pre-cluster of at least $\tau=3$ Events where one Person interacted with one Concept within $\Delta t_{\max}=30$ days.
\textbf{Task}: Synthesize a generalized Fact.

\textbf{Rules}:
\begin{itemize}
    \item \textbf{Generalization}: 3x "went hiking" $\rightarrow$ "has a habit of hiking".
    \item \textbf{Temporal Envelope}: Calculate \texttt{valid\_from} (first event) and \texttt{valid\_to} (last event).
    \item \textbf{Predicate Selection}: Choose precise verbs (e.g., \texttt{has\_habit}, \texttt{is\_experienced\_in}).
\end{itemize}

\textbf{JSON Output}:
\texttt{\{"content":"...", "predicate":"...", "valid\_from":"...", "valid\_to":"...", "confidence":0.8, "evidence\_event\_ids":[...]\}}
\end{promptbox}

\begin{promptbox}{CLAIM RECONCILIATION PROMPT (Conflict Disambiguation)}
\textbf{Input}: Claims sharing the same canonical \texttt{source}, \texttt{target}, and \texttt{predicate}, ordered by timestamp.

\textbf{Rules}:
\begin{itemize}
    \item Never merge Claims from different speakers into a consensus.
    \item Detect incompatible values or polarity from the same source and determine their temporal validity.
    \item Preserve every original Claim ID and evidence span; mark the derived view as \texttt{CURRENT}, \texttt{SUPERSEDED}, or \texttt{CONFLICTING}.
\end{itemize}

\textbf{JSON Output}:
\texttt{\{"source":"...", "target":"...", "predicate":"...", "content":"...", "valid\_from":"...", "valid\_to":"...", "status":"CURRENT", "evidence\_claim\_ids":[...]\}}
\end{promptbox}

\subsection{Answer Prompts}
\label{subsec:answer_prompts}
\begin{promptbox}{ANSWER\_PROMPT}
\textbf{Task}: Based on the retrieved evidence, generate a concise final answer.

\textbf{Refinement Rules}:
\begin{itemize}
    \item \textbf{Specific Extraction}: For "What did X research?", extract the \textit{content} (e.g., "Adoption agencies"), not the \textit{attribute} (e.g., "Research topic").
    \item \textbf{Time Formatting}: Convert ISO dates to natural language (e.g., "7 May 2023"). Prefer relative time phrases if present in the source (e.g., "the week before...").
    \item \textbf{Completeness}: List ALL items found (e.g., "violin AND clarinet").
    \item \textbf{Brevity}: Keep answers under 15 words unless explaining a complex "How/Why" question.
\end{itemize}
\end{promptbox}
\subsection{LLM-as-Judge}
\label{prompts:LLM-as-Judge}

\begin{judgepromptbox}{Temporal Reasoning Tasks (LongMemEval)}
I will give you a question, a correct answer, and a response from a model. Please answer yes if the response contains the correct answer. Otherwise, answer no. If the response is equivalent to the correct answer or contains all the intermediate steps to get the correct answer, you should also answer yes. If the response only contains a subset of the information required by the answer, answer no. In addition, do not penalize off-by-one errors for the number of days. If the question asks for the number of days/weeks/months, etc., and the model makes off-by-one errors (e.g., predicting 19 days when the answer is 18), the model's response is still correct.

\textbf{Question:} \{question\}\\
\textbf{Correct Answer:} \{answer\}\\
\textbf{Model Response:} \{response\}

Is the model response correct? Answer yes or no only.
\end{judgepromptbox}

\begin{judgepromptbox}{Standard Tasks (LongMemEval)}
I will give you a question, a correct answer, and a response from a model. Please answer yes if the response contains the correct answer. Otherwise, answer no. If the response is equivalent to the correct answer or contains all the intermediate steps to get the correct answer, you should also answer yes. If the response only contains a subset of the information required by the answer, answer no.

\textbf{Question:} \{question\}\\
\textbf{Correct Answer:} \{answer\}\\
\textbf{Model Response:} \{response\}

Is the model response correct? Answer yes or no only.
\end{judgepromptbox}

\begin{judgepromptbox}{Knowledge Update Tasks (LongMemEval)}

I will give you a question, a correct answer, and a response from a model. Please answer yes if the response contains the correct answer. Otherwise, answer no. If the response contains some previous information along with an updated answer, the response should be considered as correct as long as the updated answer is the required answer.

\textbf{Question:} \{question\}\\
\textbf{Correct Answer:} \{answer\}\\
\textbf{Model Response:} \{response\}

Is the model response correct? Answer yes or no only.
\end{judgepromptbox}

\begin{judgepromptbox}{Single-session Preference Tasks (LongMemEval)}
I will give you a question, a rubric for desired personalized response, and a response from a model. Please answer yes if the response satisfies the desired response. Otherwise, answer no. The model does not need to reflect all the points in the rubric. The response is correct as long as it recalls and utilizes the user's personal information correctly.

\textbf{Question:} \{question\}\\
\textbf{Rubric:} \{answer\}\\
\textbf{Model Response:} \{response\}

Is the model response correct? Answer yes or no only.
\end{judgepromptbox}

\begin{judgepromptbox}{Abstention Tasks (LongMemEval)}

I will give you an unanswerable question, an explanation, and a response from a model. Please answer yes if the model correctly identifies the question as unanswerable. The model could say that the information is incomplete, or some other information is given but the asked information is not.

\textbf{Question:} \{question\}\\
\textbf{Explanation:} \{answer\}\\
\textbf{Model Response:} \{response\}

Does the model correctly identify the question as unanswerable? Answer yes or no only.
\end{judgepromptbox}

\begin{judgepromptbox}{All Tasks (LoCoMo)}
\textbf{System prompt}\par
You are an expert evaluator for QA systems. 
Your task is to determine if the "Predicted Answer" is semantically CORRECT compared to the "Gold Answer".

Rules for "is\_correct":
1. TRUE if the meaning is the same, even if wording differs.
2. TRUE if the core facts are present.
3. FALSE if the answer is "I don't know" but the Gold Answer has facts.
4. FALSE if the answer points to the wrong entity, time, or location.
5. FALSE if the answer is partially correct but misses the main point.

Return JSON:
\{
    "is\_correct": <bool>,
    "explanation": "<brief explanation, max 20 words>"
\}

\textbf{User prompt}\par
Question: \{question\}\\
Gold Answer: \{gold\_answer\}\\
Predicted Answer: \{predicted\_answer\}\\
\{f'Category: \{category\}' if category else ""\}

Output the verdict in JSON format.
\end{judgepromptbox}

\end{document}